\documentclass[10pt,twocolumn,letterpaper]{article}

\usepackage{cvpr}
\usepackage{times}
\usepackage{epsfig}
\usepackage{graphicx}
\usepackage{amsmath}
\usepackage{amssymb}
\usepackage{cite}

\usepackage{booktabs} 
\usepackage{url}  
\usepackage{multirow}
\usepackage{authblk}

\usepackage{pifont}
\newcommand{\xmark}{\ding{54}}%

\usepackage[breaklinks=true,bookmarks=false]{hyperref}

\cvprfinalcopy 

\def\cvprPaperID{****} 

\ifcvprfinal\fi

\begin{document}

\title{Dual Encoding for Zero-Example Video Retrieval}


\author[1]{Jianfeng Dong}
\author[2,3]{Xirong Li\thanks{Xirong Li is the corresponding author (xirong@ruc.edu.cn).}}
\author[3]{Chaoxi Xu}
\author[4,5]{Shouling Ji}
\author[6]{Yuan He}
\author[3]{Gang Yang}
\author[1]{Xun Wang}
\affil[1]{College of Computer and Information Engineering, Zhejiang Gongshang University}
\affil[2]{Key Lab of Data Engineering and Knowledge Engineering, Renmin University of China}
\affil[3]{AI \& Media Computing Lab, School of Information, Renmin University of China}
\affil[4]{College of Computer Science, Zhejiang University}
\affil[5]{Alibaba-Zhejiang University Joint Research Institute of Frontier Technologies}
\affil[6]{Alibaba Group}

\maketitle

\begin{abstract}
This paper attacks the challenging problem of zero-example video retrieval. In such a retrieval paradigm, an end user searches for unlabeled videos by ad-hoc queries described in natural language text with no visual example provided. Given videos as sequences of frames and queries as sequences of words, an effective sequence-to-sequence cross-modal matching is required.  The majority of existing methods are concept based, extracting relevant concepts from queries and videos and accordingly establishing associations between the two modalities. In contrast, this paper takes a concept-free approach, proposing a dual deep encoding network that encodes videos and queries into powerful dense representations of their own. 
Dual encoding is conceptually simple, practically effective and end-to-end. 
As experiments on three benchmarks, \ie MSR-VTT, TRECVID 2016 and 2017 Ad-hoc Video Search show, the proposed solution establishes a new state-of-the-art for zero-example video retrieval.
\end{abstract}

\section{Introduction} \label{sec:intro}

This paper targets at \emph{zero-example} video retrieval, where a query is described in natural language text and no visual example is given. The topic is fundamentally interesting as it requires establishing proper associations between visual and linguistic information presented in temporal order.

\begin{figure}[tb!]
\centering\includegraphics[width=\columnwidth]{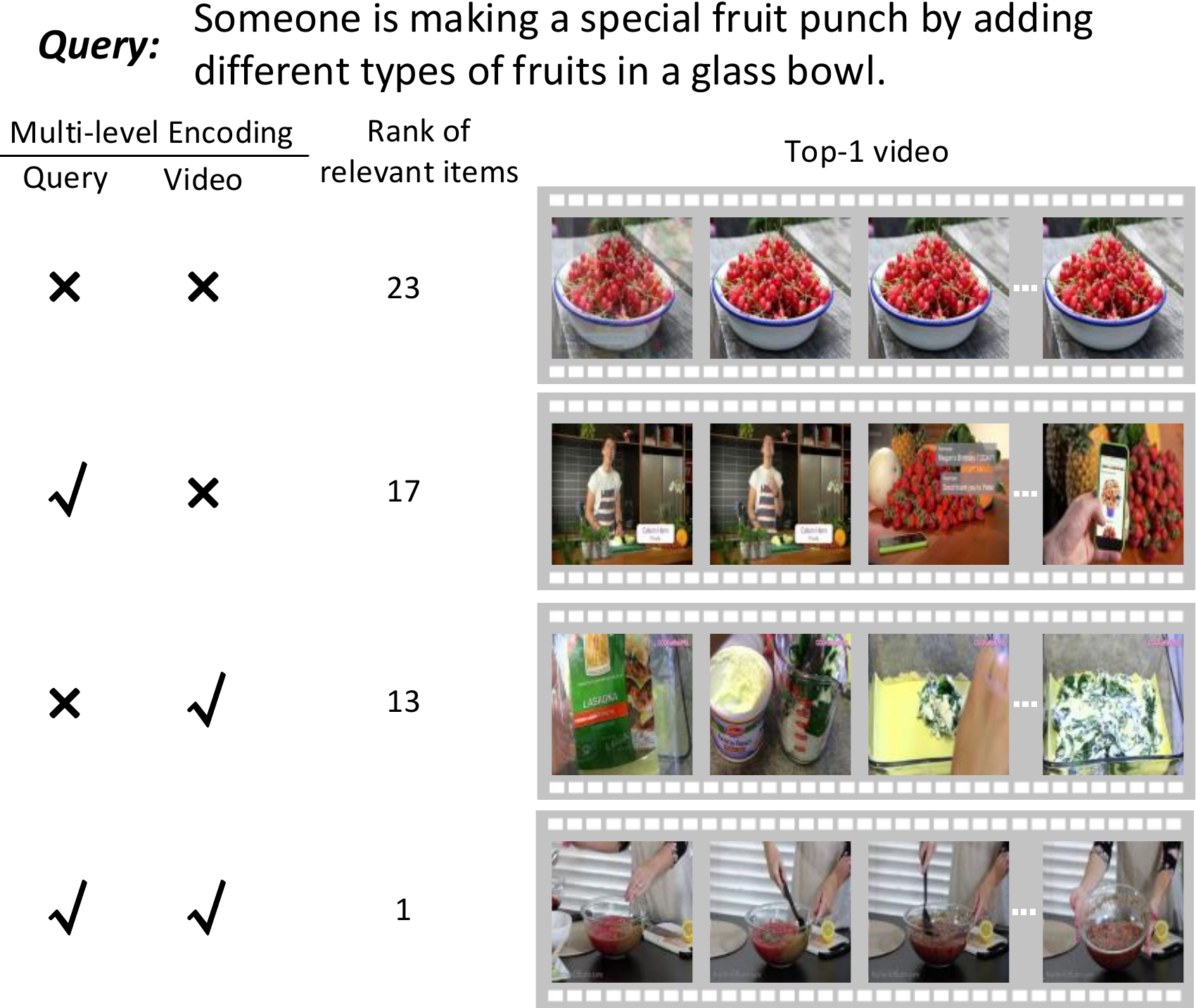}
\caption{\textbf{Showcase of zero-example video retrieval with and without the proposed encoding}. The \xmark ~symbol indicates encoding by mean pooling. Numbers in the third column are the rank of the relevant video returned by retrieval models subject to specific query / video encoding strategies. The retrieval model with dual encoding successfully answers this complex query.}\label{fig:showcase}
\end{figure}

Zero-example video retrieval attracts initial attention in the form of zero-example multimedia event detection, where the goal is to retrieve video shots showing specific events such as \emph{parking a vehicle}, \emph{dog show} and \emph{birthday party}, but with no training videos provided \cite{cikm13-zsvr,mm14-cmu-jiang,icmr14-amir,cvpr14-wu-zsed,aaai15-zsed,icmr16-zsed}. All these methods are concept based, \ie  describing the video content by automatically detected concepts, which are used to match with a target event. Such a concept-based tradition continues. For the NIST TRECVID challenge of zero-example video retrieval \cite{AwadTRECVID16}, we observe that the top performers are mostly concept based \cite{tv16-nii,tv16-certh,tv17-waseda,tv17-vireo}. However, the concept-based paradigm faces a number of difficulties including how to specify a set of concepts, how to train good classifiers for these concepts, and more crucially how to select relevant and detectable concepts for both video and query representation \cite{icmr16-zsed}. This paper differs from these works as we aim for a concept-free method that learns directly common semantic embeddings for both videos and queries.

Good efforts have been made for learning joint embeddings of the two modalities for zero-example video retrieval \cite{aaai2015-xu-video,mithun2018learning}. In \cite{aaai2015-xu-video}, a query sentence is vectorized by a recursive neural network, while \cite{mithun2018learning} vectorizes a given sentence by a recurrent neural network. In both works a specific video is vectorized by mean pooling of visual features of its frames. 
Different from \cite{aaai2015-xu-video,mithun2018learning}, we propose \emph{dual multi-level encoding} for both videos and queries in advance to common space learning. As exemplified in Figure \ref{fig:showcase}, the new encoding strategy is crucial for describing complex queries and video content. 

Our hypothesis is that a given video / query has to be first encoded into a powerful representation of its own. We consider such a decomposition crucial as it allows us to design an encoding network that jointly exploits multiple encoding strategies including mean pooling, recurrent neural networks and convolutional networks. 
In our design, the output of a specific encoding block is not only used as input of a follow-up encoding block, but also re-used via skip connections to contribute to the final output. 
It generates new, higher-level features progressively. These features, generated at distinct levels, are powerful and complementary to each other, allowing us to obtain effective video (and sentence) representations by very simple concatenation. Moreover, dual encoding is orthogonal to common space learning, allowing us to flexibly embrace  state-of-the-art common space learning algorithms.

In sum, this paper makes the following contributions. \\
$\bullet$ We propose multi-level encodings of video and text in advance to learning shared representations for the two modalities. As such, the encodings are not meant for direct video-text matching. This is conceptually different from existing works that tackle cross-modality matching as a whole.  \\
$\bullet$ We design a novel \emph{dual} network that encodes an input, let it be a query sentence or a video, in a similar manner. By jointly exploiting multi-level encodings, the network explicitly and progressively learns to represent global, local and temporal patterns in videos and sentences. While being orthogonal to common space learning, the entire model is trained in an end-to-end manner.  \\
$\bullet$ Dual encoding, combined with state-of-the-art common space learning \cite{faghri2017vse}, establishes a new state-of-the-art for zero-example video retrieval, as supported by its superior performance on three challenging benchmarks. Code and data are available at \url{https://github.com/danieljf24/dual_encoding}.

\section{Related Work} \label{sec:rel-work}

\begin{figure*}[tb!]
\centering\includegraphics[width=2\columnwidth]{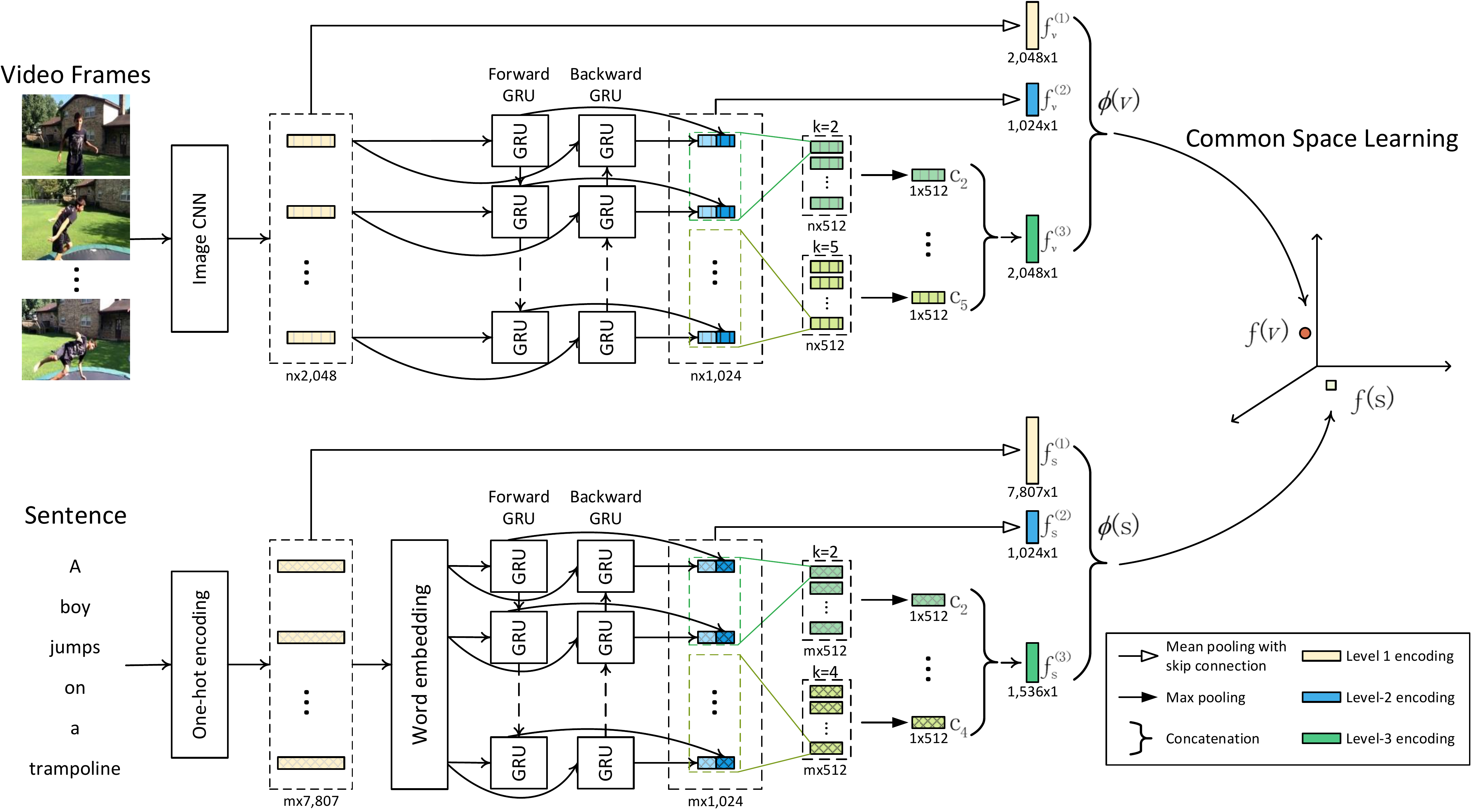}
\caption{\textbf{A conceptual diagram of the proposed dual encoding network for zero-example video retrieval}. Given a video $v$ and a sentence $s$, the network performs in parallel multi-level encodings, \ie mean pooling, biGRU and biGRU-CNN, eventually representing the two input by two combined vectors $\phi(v)$ and $\phi(s)$, respectively. The vectors are later projected into a common space, which we learn using VSE++ \cite{faghri2017vse}, for video-text similarity computation. Note that the length of the bag-of-words vector $f_s^{(1)}$ is equal to the size of the training vocabulary, which is 7,807 when we train on MSR-VTT. Once the network is trained, encoding at each side is performed independently, meaning we can process large-scale videos offline and answer ad-hoc queries on the fly.}\label{fig:framework}
\end{figure*}

Since 2016 the TRECVID starts a new challenge for zero-example video retrieval, known as Ad-hoc Video Search (AVS) \cite{AwadTRECVID16}.

The majority of the top ranked solutions for this challenge depend on visual concept classifiers to describe video content and linguistic rules to detect concept in textual queries \cite{tv16-nii,tv16-certh,icmr2017-certh-avs,tv17-waseda,tv17-vireo}.
For instance, \cite{tv16-certh,icmr2017-certh-avs} utilize multiple pre-trained Convolutional Neural Network (CNN) models to detect main objects and scenes in video frames. As for query representation, the authors design relatively complex linguistic rules to extract relevant concepts from a given query. Ueki \etal \cite{tv17-waseda} come with a much larger concept bank consisting of more than 50k concepts. In addition to pre-trained CNN models, they train SVM classifiers to automatically annotate the video content. We argue that such a concept-based paradigm has a fundamental disadvantage. That is, it is very difficult, if not impossible, to describe the rich sequential information within both video and query using a few selected concepts.

With big advances of deep learning in natural language processing and computer vision research, we observe an increased use of such techniques for video retrieval~\cite{aaai2015-xu-video,yu2017end,dong2018predicting,mithun2018learning,yu2018joint}. By directly encoding videos and text into a common space, these methods are concept free. For video encoding, a typical approach is to first extract visual features from video frames by pre-trained CNN models, and  subsequently aggregate the frame-level features into a video-level feature. To that end, mean pooling is the \emph{de facto} choice \cite{aaai2015-xu-video,dong2018predicting,mithun2018learning}. To explicitly model the temporal information, Yu \etal \cite{yu2017end} use Long Short-Term Memory (LSTM), where frame-level features are sequentially fed into LSTM, and the hidden vector at the last step is used as the video feature. CNN is exploited in \cite{yu2018joint}. None of the above works considers more than one video encoding strategy.

%
For query sentence encoding, while bag-of-words remains popular \cite{pami2017-videostory}, deep networks are in increasing use. Recursive neural networks are investigated in \cite{aaai2015-xu-video} for vectorizing subject-verb-object triplets extracted from a given sentence. Variants of recurrent neural networks are being exploited, see the usage of LSTM, bidirectional LSTM, and Gated Recurrent Unit (GRU) in \cite{yu2017end}, \cite{yu2018joint}, and \cite{mithun2018learning}, respectively. To the best of our knowledge, \cite{dong2018predicting} is the only work looking to a joint use of multiple sentence encoding strategies including bag-of-words, word2vec and GRU. However, as aforementioned, that work simply uses mean pooling for video encoding. 

To the best of our knowledge, this paper is the first work for explicitly and progressively exploiting global, local and temporal patterns in both videos and sentences.

\section{The \emph{Dual Encoding} Network} \label{sec:method}

Given a video $v$ and a sentence $s$, we propose to encode them in a dual manner, denoted as $\phi(v)$ and $\phi(s)$, in advance to common space learning. As illustrated in Figure \ref{fig:framework}, multi-level encodings are performed for each modality. The encoding results are combined to describe the two modalities in a coarse-to-fine fashion. Both video and sentence are essentially a sequence of items, let it be frames or words. Such a property allows us to design a dual encoding network to handle the two distinct modalities. In what follows we first depict the network at the video side. We then specify choices that are unique to text.

\subsection{Video-side Multi-level Encoding}

For a given video, we extract uniformly a sequence of $n$ frames with a pre-specified interval of 0.5 second. Per frame we extract deep features using a pretrained ImageNet CNN, as commonly used for video content analysis \cite{aaai2015-xu-video,pami2017-videostory,icmr2017-certh-avs}. Consequently, the video is described by a sequence of feature vectors $\{v_1,v_2,\ldots,v_n\}$, where $v_t$ indicates the deep feature vector of the $t$-th frame.  Notice that 3D CNNs \cite{tran2015learning} can also be used for feature extraction when treating segments of frames as individual items.

\subsubsection{Level 1. Global Encoding by Mean Pooling}
According to our literature review, mean pooling, which represents a video by simply averaging the features of its frames, is arguably the most popular choice for zero-example video retrieval. By definition, mean pooling captures visual patterns that repeatedly present in the video content. These patterns tend to be global. We use $\bar{v}_v$ to indicate the encoding result at this level,
\begin{equation}
f_v^{(1)} = \frac{1}{n}\sum_{t=1}^n v_t.
\end{equation}

\subsubsection{Level 2. Temporal-Aware Encoding by biGRU}
Bi-directional recurrent neural network \cite{birnn} is known to be effective for making use of both past and future contextual information of a given sequence. We hypothesize that such a network is also effective for modeling the video temporal information. We adopt a bidirectional GRU (biGRU) \cite{cho2014learning}, which has less parameters than the bidirectional LSTM and thus requires less amounts of training data.
A biGRU consists of two separated GRU layers, \ie a forward GRU and a backward GRU. The forward GRU is used to encode frame features in normal order, while the backward GRU encodes frame features in reverse order. Let $\overrightarrow{h}_t$ and $\overleftarrow{h}_t$ be their corresponding hidden states at a specific time step $t=1,\ldots,n$. The hidden states are generated as
\begin{equation}
\begin{array}{l}
\overrightarrow{h}_t = \overrightarrow{GRU}(v_t, \overrightarrow{h}_{t-1}), \\
\overleftarrow{h}_t = \overleftarrow{GRU}(v_{n+1-t}, \overleftarrow{h}_{t-1}), 
\end{array} 
\end{equation}
where $\overrightarrow{GRU}$ and $\overleftarrow{GRU}$ indicate the forward and backward GRUs, with past information carried by $\overrightarrow{h}_{t-1}$ and $\overrightarrow{h}_{t-1}$, respectively. Concatenating $\overrightarrow{h_{t}}$ and $\overleftarrow{h_{t}}$, we obtain the biGRU output $h_t = [\overrightarrow{h}_t, \overleftarrow{h}_t]$. The size of the hidden vectors in the forward and backward GRUs is empirically set to 512. Accordingly, the size of $h_t$ is 1,024. Putting all the output together, we obtain a feature map $H=\{h_1, h_2,..., h_n\}$, with a size of $1,024 \times n$. The biGRU based encoding, denoted $\bar{h}_v$, is obtained by applying mean pooling on $H$ along the row dimension, that is
\begin{equation}
f_v^{(2)} = \frac{1}{n}\sum_{t=1}^n h_t.
\end{equation}

\subsubsection{Level 3. Local-Enhanced Encoding by\\ biGRU-CNN}
The previous layer treats the output of biGRU at each step equally. To enhance local patterns that help discriminate between videos of subtle difference, we build convolutional networks on top of biGRU. In particular, we adapt 1-d CNN originally developed for sentence classification \cite{kim2014convolutional}. 

The input of our CNN is the feature map $H$ generated by the previous biGRU module. Let $Conv1d_{k,r}$ be a 1-d convolutional block that contains $r=512$ filters of size $k$, with $k \ge 2$. Feeding $H$, after zero padding, into $Conv1d_{k,r}$ produces a $n \times r$ feature map. Non-linearity is introduced by applying the ReLU activation function on the feature map. As $n$ varies for videos, we further apply max pooling to compress the feature map to a vector $c_k$ of fixed length $r$. More formally we express the above process as 
\begin{equation}
c_k = \mbox{max-pooling}(ReLU(Conv1d_{k,r}(H))).
\end{equation}

A filter with $k=2$ allows two adjacent rows in $H$ to interact with each other, while a filter of larger $k$ means more adjacent rows are exploited simultaneously. In order to generate a multi-scale representation, we deploy multiple 1-d convolutional blocks with $k=2,3,4,5$. Their output is concatenated to form the biGRU-CNN based encoding, \ie
\begin{equation}
f_v^{(3)} = [c_2, c_3, c_4, c_5].
\end{equation}

As $f_v^{(1)}$, $f_v^{(2)}$, $f_v^{(3)}$ are obtained sequentially at different levels by specific encoding strategies, we consider it reasonable to presume that the three encoding results are complementary to each other, with some redundancy. Hence, we obtain multi-level encoding of the input video by concatenating the output from all the three levels, namely
\begin{equation}
\phi(v) = [f_v^{(1)}, f_v^{(2)}, f_v^{(3)}].
\end{equation}
In fact, this concatenation operation, while being simple, is a common practice for feature combination \cite{zhou2015simple, huang2017densely}.

\subsection{Text-side Multi-level Encoding}

The above encoding network, after minor modification, is also applicable for text. 

Given a sentence $s$ of length $m$, we represent each of its words by a one-hot vector. Accordingly, a sequence of one-hot vectors $\{w_1, w_2,\ldots, w_m\}$ is generated, where $w_t$ indicates the vector of the $t$-th word. Global encoding $f_s^{(1)}$ is obtained by averaging all the individual vectors in the sequence. This amounts to the classical bag-of-words representation. 

For biGRU based encoding, each word is first converted to a dense vector by multiplying its one-hot vector with a word embedding matrix. We initialize the matrix using a word2vec \cite{word2vec} model provided by \cite{dong2018predicting}, which trained word2vec on English tags of 30 million Flickr images. The rest is mostly identical to the video counterpart. We denote the biGRU based encoding of the sentence as $f_s^{(2)}$. Similarly, we have the biGRU-CNN based encoding of the sentence as $f_s^{(3)}$. 
Here, we utilize three 1-d convolutional blocks with $k=2,3,4.$
Multi-level encoding of the sentence is obtained by concatenating the encoding results from all the three levels in the dual network, \ie
\begin{equation}
\phi(s) = [f_s^{(1)}, f_s^{(2)}, f_s^{(3)}].
\end{equation}

As $\phi(v)$ and $\phi(s)$ have not been correlated, they are not directly comparable. For video-text similarity computation, the vectors need to be projected into a common space, the learning algorithm for which will be presented next.

\section{Common Space Learning}\label{ssec:learning}

Among many choices of common space learning algorithms we choose VSE++ \cite{faghri2017vse} for two reasons. First, it is the state-of-the-art in its original context of image-text retrieval, and more recently found to be effective also in the video domain \cite{mithun2018learning}. Second, its source code is publicly available\footnote{\url{https://github.com/fartashf/vsepp}}, which greatly facilitates our exploitation of the algorithm. 

Given the encoded video vector $\phi(v)$ and sentence vector $\phi(s)$, we project them into a common space by affine transformation. From the neural network viewpoint, affine transformation is essentially a Fully Connected (FC) layer. 
On the basis of \cite{faghri2017vse}, we additionally use a Batch Normalization (BN) layer after the FC layer, as we find this trick beneficial. Putting everything together, we obtain the video feature vector $f(v)$ and sentence feature vector $f(s)$ in the common space as
\begin{equation}
\begin{array}{l}
f(v) = \mbox{BN}(W_v \phi(v) + b_v), \\
f(s) = \mbox{BN}(W_s \phi(s) + b_s), \\
\end{array} 
\end{equation}
where $W_v$ and $ W_s$ parameterize the FC layers on each side, with $b_v$ and $b_s$ as  bias terms. 

The dual encoding network and the common space learning network are trained together in an end-to-end manner except that image convnets used for extracting video features are pre-trained and fixed. Let $\theta$ be all the trainable parameters. The video-text similarity subject to $\theta$, denoted by $S_\theta(v,s)$, is computed using cosine similarity\footnote{In our preliminary experiment, we also tried the Euclidean distance, but found it less effective.} between $f(v)$ and $f(s)$. 

We use the improved marginal ranking loss \cite{faghri2017vse}, which penalizes the model according to the hardest negative examples. Concretely, the loss $\mathcal{L}(v,s; \theta)$ for a relevant video-sentence pair is defined as
\begin{equation}\label{eq:loss}
\begin{array}{r}
\mathcal{L}(v,s;\theta) = max(0, \alpha + S_\theta(v,s^-) - S_\theta(v,s)) \\
 + max(0, \alpha + S_\theta(v^-,s) - S_\theta(v,s)),
 \end{array} 
\end{equation}
where $\alpha$ is the margin constant, while $s^-$ and $v^-$ respectively indicate a negative sentence sample for $v$ and a negative video sample for $s$. The two negatives are not randomly sampled. Instead, the most similar yet negative sentence and video in the current mini-batch are chosen. The entire network is trained towards minimizing this loss.

\section{Evaluation} \label{sec:eval}

We conduct five experiments. First, following \cite{mithun2018learning}, we perform text-to-video and video-to-text retrieval on the MSR-VTT dataset \cite{xu2016msr}. We then evaluate the proposed method in the context of the TRECVID Ad-hoc Video Search task of the last two years \cite{AwadTRECVID16,AwadTRECVID17}. 
Further, we evaluate on MSVD \cite{chen2011collecting} for cross-dataset generalization and MPII-MD \cite{rohrbach2015dataset} for cross-domain generalization. While focusing on video retrieval, we provide an additional experiment on Flickr30k \cite{flickr30k} and MS-COCO \cite{lin2014microsoft} to investigate if the VSE++ model \cite{faghri2017vse}, the state-of-the-art for image-text retrieval, can be improved by replacing its GRU based encoding by the proposed encoding at the text side. Lastly, for ad-hoc video retrieval where a user submits queries on the fly, retrieval speed matters. So an efficiency test is provided.

Before proceeding to the experiments, we detail our implementations. We use PyTorch (\url{http://pytorch.org}) as our deep learning environment. For sentence preprocessing, we first convert all words to the lowercase and then replace words that occurring less than five times in the training set with a special token. We empirically set 
the size of the learned common space to 2,048, and the margin parameter $\alpha$ to 0.2. We use SGD with Adam \cite{kingma2014adam}. The mini-batch size is 128. With an initial learning rate of 0.0001, we take an adjustment schedule similar to \cite{dong2018predicting}. That is, once the validation loss does not decrease in three consecutive epochs, we divide the learning rate by 2. Early stop occurs if the validation performance does not improve in ten consecutive epochs. The maximal number of epochs is 50.

\subsection{Experiments on MSR-VTT}

\textbf{Setup}. 
The MSR-VTT dataset \cite{xu2016msr}, originally developed for video captioning, consists of 10k web video clips and 200k natural sentences describing the visual content of the clips. The average number of sentences per clip is 20. We use the official data partition, \ie 6,513 clips for training, 497 clips for validation, and the remaining 2,990 clips for testing.

For method comparison, we consider \cite{mithun2018learning}, the first work reporting video retrieval performance on MSR-VTT. A more recent work \cite{yu2018joint} also experiments with MSR-VTT, but uses a non-public subset, making its results not comparable. We include W2VV \cite{dong2018predicting}, another state-of-the-art model with code available\footnote{\url{https://github.com/danieljf24/w2vv}}. W2VV uses the Mean Square Error (MSE) loss. So for a fair comparison, we adapt the model by substituting the improved marginal ranking loss for MSE and train it from scratch. We term this variant as W2VV\textsubscript{imrl}. The same 2,048-dim ResNet-152 feature as \cite{mithun2018learning} is used.  

We report rank-based performance metrics, namely $R@K$ ($K = 1, 5, 10$), Median rank (Med r) and mean Average Precision (mAP). $R@K$ is the percentage of test queries for which at least one relevant item is found among the top-$K$ retrieved results. Med r is the median rank of the first relevant item in the search results. Higher $R@K$, mAP and lower Med r mean better performance. For overall comparison, we report the sum of all recalls. Note that for text-to-video retrieval, each test sentence is associated with one relevant video, while for video-to-text retrieval, each test video is associated with 20 relevant sentences. So the latter will have better performance scores.

\begin{table*} [tb!]
\renewcommand{\arraystretch}{1.2}
\caption{\textbf{State-of-the-art on MSR-VTT}. Larger R@\{1,5,10\}, mAP and smaller Med r indicate better performance. Methods sorted in ascending order in terms of their overall performance. The proposed method performs the best.}
\label{tab:sota-msrvtt}
\centering 
\scalebox{0.85}{
\begin{tabular}{@{}l*{12}{r}c @{}}
\toprule
\multirow{2}{*}{\textbf{Method}}   & \multicolumn{5}{c}{\textbf{Text-to-Video Retrieval}} && \multicolumn{5}{c}{\textbf{Video-to-Text Retrieval}} & \multirow{2}{*}{\textbf{Sum of Recalls}} \\
 \cmidrule{2-6}  \cmidrule{8-12} 
& R@1 & R@5 & R@10 & Med r & mAP && R@1 & R@5 & R@10 & Med r & mAP & \\
\cmidrule{1-13}
W2VV \cite{dong2018predicting}    & 1.8 & 7.0 & 10.9 & 193 & 0.052 &&      9.2 & 25.4 & 36.0 & 24 & 0.050 &     90.3 \\
Mithun \etal \cite{mithun2018learning}    & 5.8 & 17.6 & 25.2 & 61 & - &&      10.5 & 26.7 & 35.9 & 25 & - &     121.7 \\
W2VV\textsubscript{imrl}    & 6.1 & 18.7 & 27.5 & 45 & 0.131 &&      11.8 & 28.9 & 39.1 & 21 & 0.058 &     132.1 \\
\emph{Dual encoding}  & \textbf{7.7} & \textbf{22.0} & \textbf{31.8} & \textbf{32} & \textbf{0.155} &&     \textbf{13.0} & \textbf{30.8} & \textbf{43.3} & \textbf{15} & \textbf{0.065} &     \textbf{148.6}\\

\bottomrule
\end{tabular}
 }
\end{table*}

\textbf{Comparison with the State-of-the-art}.
Table \ref{tab:sota-msrvtt} shows the performance on MSR-VTT. Though our goal is zero-example video retrieval, which corresponds to text-to-video retrieval in the table, video-to-text retrieval is also included for completeness. While \cite{dong2018predicting} is less effective than \cite{mithun2018learning}, letting the former use the same loss function as the latter brings in a considerable performance gain, with the sum of recalls increased from 90.3 to 132.1. The result suggests the importance of assessing different video / text encoding strategies within the same common space learning framework. The proposed method performs the best.

\begin{table*} [tb!]
\renewcommand{\arraystretch}{1.2}
\caption{\textbf{Ablation study on MSR-VTT}. The overall performance, as indicated by \textbf{Sum of Recalls}, goes up as more encoding layers are added. Dual encoding exploiting all the three levels is the best.}
\label{tab:ablation}
\centering 
\scalebox{0.85}{
\begin{tabular}{l*{12}{r}c}
\toprule
\multirow{2}{*}{\textbf{Encoding strategy}} & \multicolumn{5}{c}{\textbf{Text-to-Video Retrieval}} && \multicolumn{5}{c}{\textbf{Video-to-Text Retrieval}} &  \multirow{2}{*}{\textbf{Sum of Recalls}}\\
\cmidrule(l){2-6} \cmidrule(l){8-12}
  &  R@1 & R@5 & R@10  & Med r & mAP && R@1 & R@5 & R@10  & Med r & mAP & \\
\cmidrule(l){1-13}
Level 1 (Mean pooling)    &  6.4 & 18.8 & 27.3 & 47 & 0.132  &&     11.5 & 27.7 & 38.2 & 22 & 0.054 &     124.4 \\
Level 2 (biGRU)          &  6.3 & 19.4 & 28.5 & 38 & 0.136 &&     10.1 & 26.8 & 37.7 & 20 & 0.057 &      124.8 \\
Level 3 (biGRU-CNN)      &  7.3 & 21.5 & 31.2 & 32 & 0.150  &&     10.6 & 27.3 & 38.5 & 20 & 0.061 &      136.4 \\
Level 1 + 2               &  6.9 & 20.4 & 29.1 & 41 & 0.142  &&     11.6 & 29.6 & 40.7 & 18 &  0.058 &     138.3 \\
Level 1 + 3               &  7.5 & 21.6 & 31.2 & 33 & 0.151  &&     11.9 & 30.5 & 41.7 & 16 & 0.062 &      144.7 \\
Level 2 + 3               &  7.6 & \textbf{22.4} & \textbf{32.2}  & \textbf{31} & \textbf{0.155}  &&      11.9 & \textbf{30.9} & 42.7& 16 & \textbf{0.066} &      147.6 \\ 
Level 1 + 2 + 3           & \textbf{7.7} & 22.0 & 31.8 & 32 & \textbf{0.155}  &&       \textbf{13.0} & 30.8 & \textbf{43.3} & \textbf{15} & 0.065 &    \textbf{148.6} \\
\bottomrule
\end{tabular}
 }
\end{table*}

\textbf{Ablation Study}. To exam the usefulness of each component in the dual encoding network, we conduct an ablation study as follows. Given varied combinations of the components, seven models are trained. Table \ref{tab:ablation} summarizes the choices of video and text encodings and the corresponding performance.
Among the individual encoding levels, biGRU-CNN, which builds CNN on top of the output of biGRU turns out to be the most effective. 
As more encoding layers are included, the overall performance goes up. For the last four models which combines output from previous layers, they all outperform the first three models. This suggests that different layers are complementary to each other. The full multi-level encoding setup, \ie Level 1 +2 + 3 in Table \ref{tab:ablation}, is the best.

We also investigate single-side encoding, that is, video-side multi-level encoding with mean pooling on the text side and text-side multi-level encoding with mean pooling on the video side. These two strategies obtain Sum of Recalls of 143.6 and 137.1, respectively. The lower scores justify the necessity of dual encoding. The result also suggests that video-side encoding is more beneficial. 

\subsection{Experiments on TRECVID}

\begin{table} [tb!]
\renewcommand{\arraystretch}{1.2}
\caption{\textbf{State-of-the-art on TRECVID 2016}. }
\label{tab:tv16-avs}
\centering 
\scalebox{0.85}{
\begin{tabular}{@{} lr @{}}
\toprule
\textbf{Method} & \textbf{infAP} \\
\cmidrule{1-2} 

\emph{Top-3 TRECVID finalists}: \\ 
Le \etal \cite{tv16-nii}   & 0.054 \\
Markatopoulou \etal \cite{tv16-certh} & 0.051 \\
Liang \etal \cite{tv16-inf}   & 0.040 \\  [3pt]

\emph{Literature methods}: \\
Habibian \etal \cite{pami2017-videostory} &  0.087 \\ 
Markatopoulou \etal \cite{icmr2017-certh-avs}  &  0.064 \\ 

\cmidrule{1-1}
W2VV\textsubscript{imrl}  & 0.132 \\
\emph{Dual encoding} & \textbf{0.159}  \\ 
\bottomrule
\end{tabular}
}
\end{table}

\begin{table} [tb!]
\renewcommand{\arraystretch}{1.2}
\caption{\textbf{State-of-the-art on TRECVID 2017}. }
\label{tab:tv17-avs}
\centering 
\scalebox{0.85}{
\begin{tabular}{@{} lr @{}}
\toprule
\textbf{Method} & \textbf{infAP} \\
\cmidrule{1-2} 

\emph{Top-3 TRECVID finalists}: \\ 
Snoek \etal \cite{tv17-uva}    & 0.206 \\
Ueki \etal \cite{tv17-waseda} & 0.159 \\
Nguyen \etal \cite{tv17-vireo}  & 0.120 \\  [3pt]
\emph{Literature methods}: \\
Habibian \etal \cite{pami2017-videostory} & 0.150  \\ 

\cmidrule{1-1}
W2VV\textsubscript{imrl}  & 0.165 \\
\emph{Dual encoding}  & \textbf{0.208}   \\ 
\bottomrule
\end{tabular}
}
\end{table}

\textbf{Setup}. 
We evaluate dual encoding in the TRECVID AVS task \cite{AwadTRECVID16,AwadTRECVID17}, which provides the largest test bed for zero-example video retrieval to this date. The test collection, called IACC.3, contains 4,593 Internet Archive videos with duration ranging from 6.5 min to 9.5 min and a mean duration of almost 7.8 min. Shot boundary detection results in 335,944 shots in total. Given an ad-hoc query, \eg \emph{Find shots of military personnel interacting with protesters}, the task is to return for the query a list of 1,000 shots from the test collection ranked according to their likelihood of containing the given query. Per year TRECVID specifies 30 distinct queries of varied complexity.

As TRECVID does not specify training data for the AVS task, we train the dual encoding network using the joint collection of MSR-VTT and the TGIF \cite{tgif}, which contains 100K animated GIFs and 120K sentences describing visual content of the GIFs. Although animated GIFs are a very different domain, TGIF was constructed in a way to resemble user-generated video clips, \eg with cartoon, static, and textual content removed. For IACC.3, MSR-VTT and TGIF, we use frame-level CNN features provided by \cite{tv18-ruc}, where the authors use ResNeXt-101 \cite{resnext} trained on the full ImageNet collection for feature extraction. 

For method comparison, we include the top 3 entries of each year, \ie \cite{tv16-nii,tv16-certh,tv16-inf} for 2016 and \cite{tv17-uva,tv17-waseda,tv17-vireo} for 2017. Besides we include publications that report performance on the tasks, \ie \cite{pami2017-videostory,icmr2017-certh-avs}, to the best of our knowledge. As noted in our literature review, most of the methods are concept based. Notice that visual features and training data used by these methods vary, meaning the comparison and consequently conclusions drawn from this comparison is at a system level. So for a more conclusive comparison, we re-train W2VV\textsubscript{imrl} using the same joint dataset and the same ResNeXt-101 feature.

We report inferred Average Precision (infAP), the official performance metric used by the TRECVID AVS task. The overall performance is measured by averaging infAP scores over the queries.

\textbf{Comparison with the State-of-the-art}
Table \ref{tab:tv16-avs} and \ref{tab:tv17-avs} show the performance of different methods on the TRECVID 2016 and 2017 AVS tasks, respectively. The proposed method again performs the best, with infAP of 0.159 and 0.208. While \cite{tv17-uva} has a close infAP of 0.206 on the TRECVID 2017 task, their solution ensembles ten models. Their best single model, \ie \cite{pami2017-videostory} which uses the same ResNeXt-101 feature, has a lower infAP of 0.150. Given the same training data and feature, the proposed method outperforms W2VV\textsubscript{imrl} as well. Table \ref{tab:avs_train_data} shows the influence of distinct training data. The above results again justify the effectiveness of dual encoding.

\begin{figure*}[tb!]
\centering\includegraphics[width=2\columnwidth]{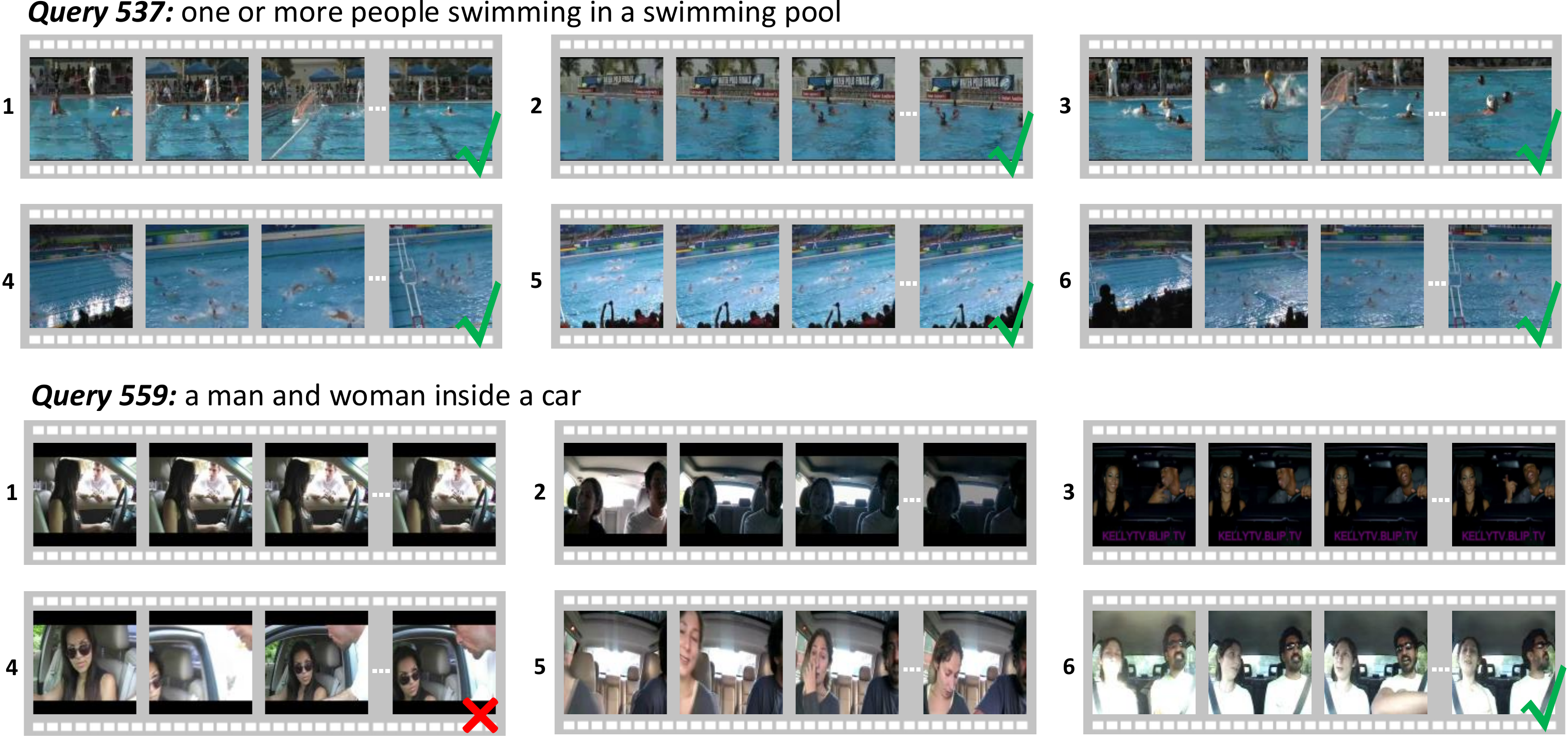}
\caption{\textbf{Top six shots retrieved from the IACC.3 collection (of 335k shots) with respect to four queries in the TRECVID 2017 AVS task}. Dual encoding is used. Videos without green or red marks are unlabeled. For query \#559, the second, third and fifth shots are unlabeled but seem to be relevant to the query. The fourth shot is incorrect, as our video retrieval model fails to recognize that the man is in fact outside a car.}\label{fig:tv17avs-showcase}
\end{figure*}

Note that the TRECVID ground truth is partially available at the shot-level. The task organizers employ a pooling strategy to collect the ground truth, \ie a pool of candidate shots are formed by collecting the top-1000 shots from each submission and a random subset is selected for manual verification. The ground truth thus favors official participants.
As the top ranked items found by our method can be outside of the subset, infAP scores of our method is likely to be underestimated. Some qualitative results are show in Fig. \ref{fig:tv17avs-showcase}.

\begin{table} [tb!]
\renewcommand{\arraystretch}{1.2}
\caption{\textbf{Performance of our model trained on distinct data for the TRECVID AVS benchmarks}. Performance metric: infAP.}
\label{tab:avs_train_data}
\centering
 \scalebox{0.8}{
     \begin{tabular}{@{} l rrr @{}}
\toprule
\textbf{Training data}  & \textbf{TRECVID 2016}   & \textbf{TRECVID 2017} \\
\midrule
    MSR-VTT           & 0.140 & 0.179 \\
    TGIF              & 0.092 & 0.199 \\
    MSR-VTT + TGIF    & \textbf{0.159} & \textbf{0.208}  \\
\bottomrule
\end{tabular}
 }
\end{table}

\subsection{Experiments on MSVD and MPII-MD}

\begin{table} [tb!]
\renewcommand{\arraystretch}{1.2}
\caption{\textbf{Performance of zero-example video retrieval, measured by mAP}. Our proposed method is the best.}
\label{tab:msvd_mpii}
\centering
 \scalebox{0.85}{
     \begin{tabular}{@{} l rr @{}}
\toprule
\textbf{Model}   & \textbf{MSVD} & \textbf{MPII-MD} \\
\midrule 
    W2VV                         & 0.100  & 0.008  \\
    W2VV\textsubscript{imrl}     & 0.230  & 0.030  \\
    VSE++                        & 0.218  & 0.022  \\
    \emph{Dual Encoding}         & \textbf{0.232} & \textbf{0.037}  \\
\bottomrule
\end{tabular}  
 }
\end{table}

\textbf{Setup}. We evaluate on MSVD \cite{chen2011collecting} and MPII-MP \cite{rohrbach2015dataset}, two other popular video sets. Note that MSR-VTT is built in a similar vein to MSVD , but in a larger scale. 
So we assess the models previously trained on MSR-VTT using the MSVD test set.
MPII-MD, as a movie description dataset, is unique. 
So we re-train and evaluate all the models on this dataset with its official data split. The ResNeXt-101 feature is used.

\begin{figure}[tb!]
\centering
\includegraphics[width=\columnwidth]{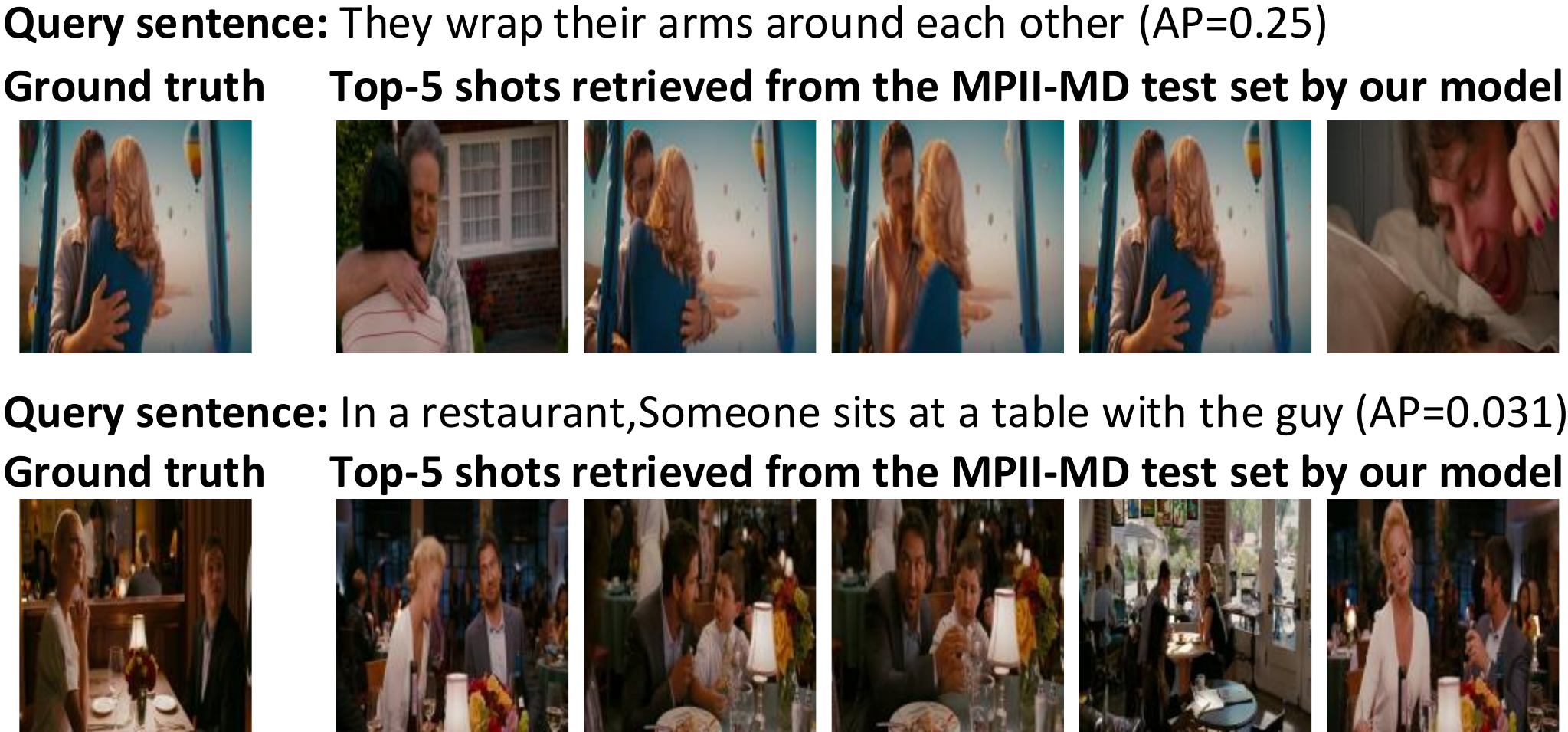}
\caption{\textbf{Movie retrieval by sentence on MPII-MD}. The top retrieved shots, though not being ground truth, appear to be correct. }
\label{fig:mpii-md}
\end{figure}

\textbf{Results}. 
As Table \ref{tab:msvd_mpii} shows, our model again performs the best in the cross-dataset scenario. Our model is also the most effective on MPII-MP. See Fig. \ref{fig:mpii-md} for qualitative results of zero-example movie retrieval. 


\subsection{Experiments on Flickr30K and MSCOCO}

\textbf{Setup}.
We investigate if the VSE++ model \cite{faghri2017vse} can be improved in its original context of image-text retrieval, when replacing its textual encoding module, which is a GRU, by the proposed multi-level encoding module. To that end, we fix all other choices, adopting the exact evaluation protocol of \cite{faghri2017vse}. That is, we use the same data split, where the training / validation / test test has 30,000 / 1,000 / 1,000 images for Flickr30K, and 82,783 / 5,000 / 5,000 images for MSCOCO. We also use the same VGGNet feature provided by \cite{faghri2017vse}. Performance of $R@1$, $R@5$ and $R@10$ are reported. On MSCOCO, the results are reported by averaging over 5 folds of 1,000 test images.


\begin{table} [tb!]
\renewcommand{\arraystretch}{1.2}
\caption{\textbf{Performance of image-text retrieval on Flickr30k and MSCOCO}. The proposed multi-level encoding is beneficial for the VSE++ model \cite{faghri2017vse}.}
\label{tab:image-text}
\centering 
\scalebox{0.72}{
\begin{tabular}{@{}l*{7}{r} @{}}
\toprule
\multirow{2}{*}{\textbf{Method}}   & \multicolumn{3}{c}{\textbf{Text-to-Image}} && \multicolumn{3}{c}{\textbf{Image-to-Text}} \\
 \cmidrule{2-4}  \cmidrule{6-8} 
& R@1 & R@5 & R@10 && R@1 & R@5 & R@10  \\
\cmidrule{1-8}
\textit{On Flickr30k}                       &&& & &  &    &   \\
VSE++  & 23.1 & 49.2 & 60.7 &&     31.9 & 58.4 & 68.0  \\
VSE++, multi-level encoding  & \textbf{24.7} & \textbf{52.3} & \textbf{65.1} &&      \textbf{35.1} & \textbf{62.2} & \textbf{71.3}  \\
\cmidrule{1-8}
\textit{On MSCOCO}                         &&&  & &  &    &   \\
VSE++  & 33.7 & 68.8 & 81.0 && 43.6 & 74.8 & 84.6       \\
VSE++, multi-level encoding  &  \textbf{34.8} & \textbf{69.6} & \textbf{82.6} &&   \textbf{46.7} & \textbf{76.2} & \textbf{85.8}    \\
\bottomrule
\end{tabular}
 }
\end{table}

\textbf{Results}.
Table \ref{tab:image-text} shows the performance of image-text retrieval on Flickr30k and MSCOCO. Integrating text-side multi-level encoding into VSE++ brings improvements on both datasets.  The results suggest that the proposed text-side multi-level encoding is also beneficial for VSE++ in its original context.

\subsection{Efficiency Test}

Recall that the dual encoding network is designed to represent both videos and sentences into a common space respectively. Once the network is trained, representing them in the common space can be performed independently. This means we can process large-scale videos offline and answer ad-hoc queries on the fly. 
Specifically, given a natural-sentence query, it takes approximately 0.14 second to retrieve videos from the largest IACC.3 dataset, which consists of 335,944 videos. The performance is tested on a normal computer with 64G RAM and a GTX 1080TI GPU. The retrieval speed is adequate for instant response.

\section{Summary and Conclusions} \label{sec:conc}

For zero-example video retrieval this paper proposes dual encoding. By jointly exploiting multiple encoding strategies at different levels, the proposed dual encoding network encodes both videos and natural language queries into powerful dense representations. Followed by common space learning, these representations can be transformed to perform sequence-to-sequence cross-modal matching effectively.  Extensive experiments on three benchmarks, \ie MSR-VTT, TRECVID 2016 and 2017 AVS tasks, support the following conclusions. Among the three levels of encoding, biGRU-CNN  that builds a 1-d convolutional network on top of bidirectional GRU is the most effective when used alone. Video-side multi-level encoding is more beneficial when compared with its text-side counterpart. For state-of-the-art performance, we recommend dual encoding.  
We believe the proposed method also has a potential for other tasks such as video question answering that require effective video / text encoding.

\section*{Acknowledgments}
This work was supported by NSFC (No. 61672523, No. 61773385, No. U1609215, No. 61772466), ZJNSF (No. LQ19F020002), the Fundamental Research Funds for the Central Universities and the Research Funds of Renmin University of China (No. 18XNLG19),  and the Zhejiang Provincial Natural Science Foundation for Distinguished Young Scholars (No. LR19F020003).


{\small
\bibliographystyle{ieee}
\bibliography{egbib}
}

\end{document}